\documentclass[10pt,conference]{IEEEtran}

\usepackage{fancyhdr, etoolbox}
\fancypagestyle{firstpage}{%
    \fancyhead[L]{}
    \fancyhead[C]{The 1st International Workshop on Interpretability and Robustness in Neural Software Engineering (InteNSE'23)}
    \fancyhead[R]{}
}

\usepackage[numbers, sort]{natbib}
\usepackage{comment}
\usepackage{balance}
\usepackage{amsmath}
\usepackage{hyperref}
\usepackage{cleveref}
\usepackage{mathtools}
\usepackage{multicol}
\usepackage{multirow}
\usepackage{xspace}
\usepackage{xcolor}
\usepackage{stfloats}
\usepackage{fancybox}
\usepackage{graphicx}
\usepackage{lipsum}
\usepackage{enumitem}
\usepackage{pbox}
\usepackage{float}
\usepackage{url} 
\usepackage{array}
\usepackage{svg}
\usepackage[normalem]{ulem}
\usepackage[labelfont={bf}, font={normalsize}]{caption}
\usepackage{subcaption}
\newcommand{\eg}{e.g.\xspace}
\newcommand{\ie}{i.e.\xspace}
\newcommand{\etal}{\emph{et al.}\xspace}

\newcommand{\aka}{\emph{a.k.a.}\xspace}

\newcommand{\Part}[1]{\noindent\textbf{#1}}

\newcommand{\Space}[1]{}

\newcolumntype{C}[1]{>{\centering\let\newline\\\arraybackslash\hspace{0pt}}m{#1}}

\newcommand{\DD}{Delta Debugging\xspace}
\newcommand{\ddmin}{\textit{ddmin}\xspace}

\newcommand{\ci}{code intelligence\xspace}

\newcommand{\mn}{\textsc{MethodNaming}\xspace}
\newcommand{\vm}{\textsc{VarMisuse}\xspace}
\newcommand{\vd}{\textsc{VulDetection}\xspace}
\newcommand{\cs}{\textsc{CodeSearch}\xspace}

\newcommand{\ctv}{Code2Vec\xspace}
\newcommand{\cts}{Code2Seq\xspace}
\newcommand{\tra}{Transformer\xspace}
\newcommand{\rnn}{RNN\xspace}
\newcommand{\cnn}{CNN\xspace}
\newcommand{\cbert}{CodeBERT\xspace}

\newcommand{\jlarge}{Java-Large\xspace}
\newcommand{\pyg}{Py150-Great\xspace}

\newcommand{\sbabi}{s-bAbI\xspace}
\newcommand{\csn}{CodeSearchNet\xspace}
\newcommand{\csnjava}{CSN-Java\xspace}

\begin{document}

\title{Study of Distractors in Neural Models of Code}

\author{
    \IEEEauthorblockN{Md Rafiqul Islam Rabin}
    \IEEEauthorblockA{
        \textit{mrabin@uh.edu} \\
        University of Houston \\
        Houston, TX, USA
    }
    \and
    \IEEEauthorblockN{Aftab Hussain}
    \IEEEauthorblockA{
        \textit{ahussain27@uh.edu} \\
        University of Houston \\
        Houston, TX, USA
    }
    \and
    \IEEEauthorblockN{Sahil Suneja}
    \IEEEauthorblockA{
        \textit{suneja@us.ibm.com} \\
        IBM Research \\
        Yorktown Heights, NY, USA
    }
    \and
    \IEEEauthorblockN{Mohammad Amin Alipour}
    \IEEEauthorblockA{
        \textit{maalipou@central.uh.edu} \\
        University of Houston \\
        Houston, TX, USA
    }
}

\maketitle
\thispagestyle{firstpage}

\begin{abstract}
Finding important features that contribute to the prediction of neural models is an active area of research in explainable AI. 
Neural models are opaque and finding such features sheds light on a better understanding of their predictions. 
In contrast, in this work, we present an inverse perspective of \emph{distractor features}: features that cast doubt about the prediction by affecting the model's confidence in its prediction.
Understanding distractors provide a complementary view of the features' relevance in the predictions of neural models. 

In this paper, we apply a reduction-based technique to find distractors and provide our preliminary results of their impacts and types. 
Our experiments across various tasks, models, and datasets of code reveal that the removal of tokens can have a significant impact on the confidence of models in their predictions and the categories of tokens can also play a vital role in the model's confidence.
Our study aims to enhance the transparency of models by emphasizing those tokens that significantly influence the confidence of the models.
\end{abstract}


 \section{Introduction}

Deep learning has been increasingly used in software tools to analyze existing programming data and produce new insights. The application of deep learning ranges from discriminative models that predict properties of programs \cite{allamanis2018survey}, \eg, type prediction, bug detection, and clone detection, to more complex generative language models that output strings of characters \cite{CodeXGLUE}, \eg, code generation, code summarization, and code search.
Despite the recent strides in applications of deep learning models for various software engineering tasks, these models are still opaque and there is still little known about why they make a certain prediction and what input features influence them, thereby hindering their trustworthiness.

Recent studies have shown that neural models for \ci highly depend on the uncurated, sometimes duplicate data used for training \cite{allamanis2019duplication, rabin2023memorization} and usually suffer from generalization performance with real-world data beyond known training scenarios \cite{kang2019generalizability, rabin2021generalizability}.
To uncover model learning, a few studies have investigated the area of explainability for \ci models, using techniques such as feature selection \cite{allamanis2015suggesting, rabin2020demystifying}, code perturbation and mutation \cite{bui2019autofocus, wang2022demystifying, cito2022counterfactual}, and program simplification \cite{rabin2021dd, suneja2021probing}, amongst others.
These works aim to detect subsets of the most important input features that positively contribute to the models' predictions.
In contrast, in this work, we focus on the inverse problem: \emph{what features cast doubt about a prediction, or potentially hinder a model's confidence?}
More precisely, in the domain of \ci, we define the notion of \emph{distractor} tokens, \ie, the \emph{confidence} of the model in its prediction changes significantly when such tokens are removed from the input programs.
The confidence of the model, which is often represented as a probability score, indicates how certain a model is in the prediction.
Therefore, similar to important features in input programs, but unlike existing works, learning about distractors that influence the model's confidence, would also allow better interpretation and representations in neural models of code.

\begin{figure*}[htbp]
  \centering
  \includegraphics[width=0.92\textwidth]{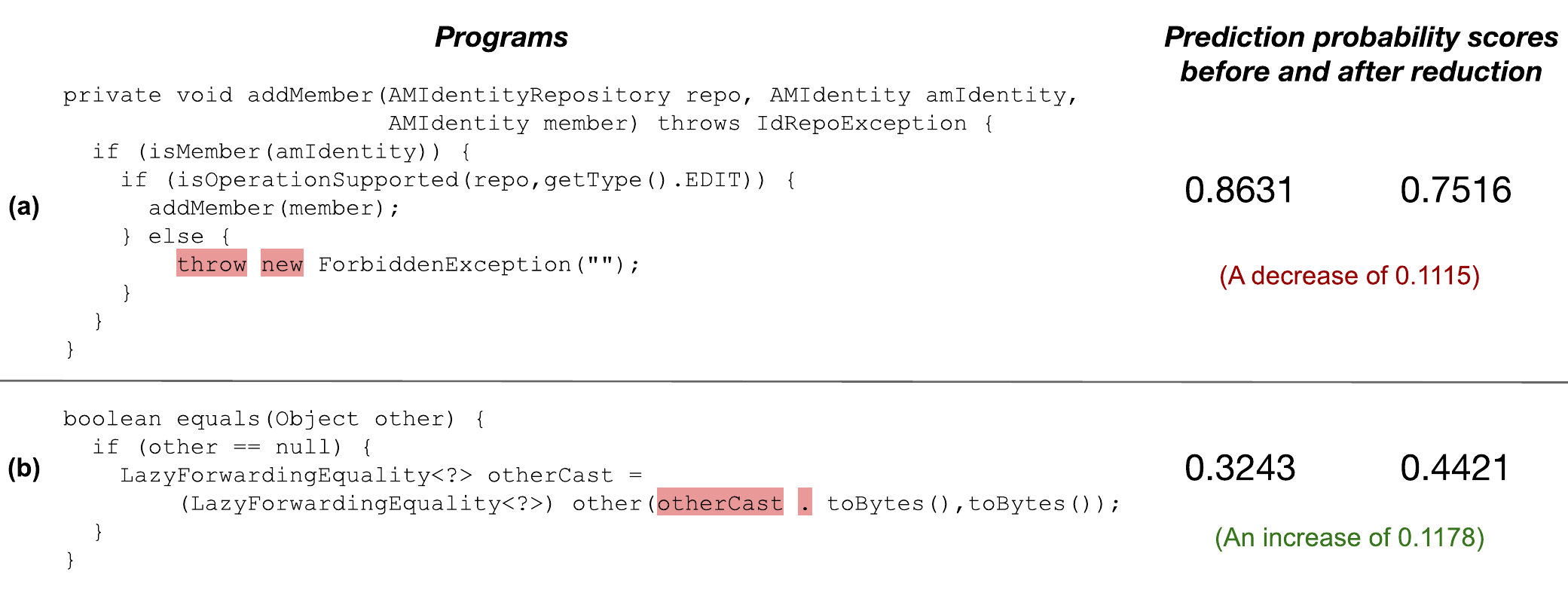}
  \caption{Effect of removing tokens on the probability score of a model's prediction. Here, we provide two programs that show \cts's corresponding prediction probability scores before and after removing tokens (highlighted in red) for the method naming task.
  Verification of the model's reliance on these tokens vs. the ground truth can help reveal the model learning quality.}
  \label{fig:skepticism_example}
\end{figure*}

\Cref{fig:skepticism_example} illustrates the distractors' influence on the confidence of \cts model \cite{alon2019code2seq}, as a motivating example.
Omitting the keyword tokens \texttt{throw} and \texttt{new} from an input program significantly reduces the probability score of the model's prediction, indicating a drop in the model's confidence during prediction.
Similarly, omitting the variable-name token \texttt{otherCast}, and the member-access dot operator, contributes to an increase in the prediction probability score.
This example demonstrates that some tokens can have a large effect on the model's confidence. 
To understand this impact better, it is necessary to study how different tokens affect the model’s prediction. This research takes a first step in that direction by looking at the influence of various tokens on a model’s prediction. 
By understanding which tokens have a positive or negative impact on the confidence of the model, we can adjust data pre-processing and normalization to assist the model in focusing on more task-relevant code constructs, thereby improving the model's reliability and trustworthiness.

In this study, we propose an approach to identify and evaluate the extent of distractors in \ci models. Our approach was built on state-of-the-art program reduction frameworks (SIVAND \cite{rabin2021dd,rabin2022perses} and P2IM \cite{suneja2021probing,suneja2021enhancement}) for \ci. It uses an iterative process for simplifying input programs using the \DD \cite{zeller2002dd} algorithm to find the most relevant input features. It analyzes the output of each reduction step and identifies the tokens whose reduction significantly increases or decreases the confidence of the models in its prediction.

Our experiments with seven \ci models and four datasets across four tasks reveal that the \ctv model of \mn task is comparatively more susceptible to distractors and the \cbert model of \cs task has a comparatively higher reliance on individual tokens.
Furthermore, our results highlight that some categories of tokens, \eg, control structures, may play a significant role in the model’s confidence. 
Although we focus on the popular study subjects of existing explainability research, our approach is applicable to other model architectures, downstream tasks, and target datasets.
This paper makes the following contributions.
\begin{itemize}
    \item We introduce the concepts of distractors in the neural models of code.
    \item We evaluate the extent of distractors on several tasks, models, and datasets of code.
    \item We discuss the implication of our findings on the debugging of neural models of code.
\end{itemize}

\section{Methodology}
\label{sec:methodology}

This section describes our approach for extracting distractor tokens in neural \ci models. To understand the impact of individual tokens on the model's confidence during prediction, we remove tokens from programs systematically using a program reduction algorithm while preserving the prediction of the model. We next study the changes in the confidence of the model in reduced programs for observing the impact of removed tokens.


\medskip
\Part{\DD}.
Zeller \etal \cite{zeller2002dd} have proposed the \DD (\ddmin) algorithm for reducing the size of a failure-inducing input program. The goal of this algorithm is to identify a minimal program that preserves the same behavior of the input program. For instance, given a failure-inducing input program, the \ddmin algorithm systematically removes tokens from the input program while preserving the failure.
It mainly follows the steps below:
\begin{itemize}
    \item It first splits the input into a sequence of smaller chunks which are called \textit{deltas}. Next, it iteratively creates and evaluates candidates by removing deltas from the input.
    \item The algorithm then checks if any resulting candidate preserves the same behavior of the input, \eg, failure. If yes, it uses the candidate as the new base to be reduced further. Otherwise, it increases the granularity for deltas, until it determines that the input cannot be reduced further.
    \item The final reduced candidate is \textit{1-minimal}, where any further attempts to reduce it would change the behavior of the input.
\end{itemize}

Rabin \etal \cite{rabin2021dd,rabin2022perses} and Suneja \etal \cite{suneja2021probing,suneja2021enhancement} have applied \DD to enhance the transparency of neural \ci models by finding the most relevant features to the model's prediction. In their approaches, they use the \ddmin algorithm to reduce the size of input programs while preserving the predictions of models.
Suppose, $y = M(p_o)$ denotes that the output of model $M$ on the input program $p_o$ is $y$. They use \ddmin to reduce $p_o$ to $p_r$, where size of $p_r$ is substantially smaller than $p_o$, \ie, $|p_r| < |p_o|$, and $p_r$ is 1-minimal.
In their approaches, they contrast the original input program $p_o$ and its corresponding 1-minimal reduced program $p_r$ to identify the relevant features in $p_o$ that contribute the most to the prediction of the model, \ie, $M(p_o)$.


\medskip
\Part{Distractor Tokens}.
While almost all prior studies mainly focus on understanding key input features that are most relevant to the models' prediction, this work is concerned with the impact of tokens on the confidence of models in their prediction. While the prediction may stay the same, removing some tokens may change the confidence of the model during prediction.
To this end, in the consecutive steps where the prediction of the model remains unchanged, we check if (1) any tokens have been removed, and (2) the confidence of the model has been impacted noticeably. In this work, we consider an impact noticeable, if the probability score of a model's prediction changes by at least $0.1$, that is $10\%$.

After each step of token removal from the input program using \ddmin, we check the reduced program to observe if the removal has any noticeable impact on the model's prediction. If the prediction probability score increases/decreases and the target prediction label remains unchanged, we consider this as a drastic probability change.
We then collect those removed tokens as potential \emph{distractor} tokens for target prediction.

We adapt the reduction steps of \ddmin as follows:
\begin{itemize}
    \item Given an input program $p_o$ and a \ci model $M$, we record the predicted label $y_o$ and probability score $s_o$ given by the model $M$ on the input $p_o$.
    
    \[ y_o, s_o = M(p_o) \]
    
    \item Using \ddmin, we then produce a reduced program $p_r$ by removing some tokens $t_r$ from the input program $p_o$ while preserving the same predicted label $y_o$. Next, we compute the predicted label $y_r$ and probability score $s_r$ by the model $M$ on the reduced program $p_r$.
    
    \begin{align*}
    p_r, t_r &= ddmin(p_o) \\
    y_r, s_r &= M(p_r)
    \end{align*}
    
    \item Finally, we compare the probability score $s_o$ of the input program with the probability score $s_r$ of the reduced program for understanding the impact of tokens $t_r$. If removing tokens significantly changes the model's probability score ($\Delta{s} = | s_r - s_0 |$), then those tokens are identified as distractor tokens.
    
    \begin{align*}
    \text{distractor tokens}, t_r = \Biggl \{ {{s_r \gg s_0 : +\Delta{s}} \atop {s_r \ll s_0 : -\Delta{s}}}
    \end{align*}
\end{itemize}

As \ddmin algorithm reduces an original input program ($p_o$) to the final 1-minimal program ($p_r$), it generates several intermediate reduced programs that preserve the original prediction ($M(p_o)$). 
Suppose, there are $n$ such intermediate reduced programs to get the final reduced program for a sample. For each input program, we then compute an average probability change as $\frac{1}{n} \sum {(\Delta{s_1}+...+\Delta{s_i}+...+\Delta{s_n})}$ and maximum probability change as $\max(\Delta{s_1}, ..., \Delta{s_i}, ..., \Delta{s_n})$, where $\Delta{s_i} = | s_{i} - s_{i-1} |$ and $s_{i}$ is the prediction probability score given by a model on the reduced program of $i^{th}$ step.

\section{Study Subjects}
\label{sec:settings}

In this section, we provide a brief description of the design and implementation aspects of different tasks, models, and datasets used in our study. We study four well-known tasks (software vulnerability detection, method name prediction, variable misuse localization-repair, and natural language code search) across seven models (\rnn, \cnn, \ctv, \cts, \rnn, \tra, and \cbert) and four datasets (\sbabi, \jlarge, \pyg, and \csnjava) of three programming languages (C, Java, and Python).

The software vulnerability detection \cite{suneja2021probing, suneja2021enhancement}, method name prediction \cite{allamanis2015suggesting, allamanis2016summarization}, variable misuse localization-repair \cite{allamanis2018ggnn, vasic2019neural}, and natural language code search \cite{feng2020codebert, CodeXGLUE} tasks have been heavily studied for the evaluation and transparency of neural \ci models \cite{wang2019coset, kang2019generalizability, rabin2019tnpa, compton2020obfuscation, rabin2021generalizability, yefet2020adversarial, rabin2020demystifying, rabin2021code2snapshot, rabin2021dd, rabin2022perses, suneja2021probing, suneja2021enhancement, wang2022demystifying, rabin2023memorization}. For these tasks, we choose popular models and datasets for which the training artifacts are publicly available \cite{alon2019code2seq,hellendoorn2020global,suneja2021probing,feng2020codebert}. We use the default base configurations given by the authors in their paper or repository, to which we only make minor modifications, such as changing the batch size.

\subsection{Software Vulnerability Detection (\vd)}

\Part{Task:}
In the software vulnerability detection task, the goal is to detect potential vulnerabilities (\aka bugs or defects) in source code. Models are trained with respect to samples indicating whether or not a given piece of code contains vulnerabilities and then used to predict whether new pieces of code are likely to contain any vulnerabilities. This task has been used as the downstream task for different applications such as probing \cite{suneja2021probing}, understanding \cite{suneja2021reliable}, and enhancing \cite{suneja2021enhancement} models' signal awareness and reliability issues.

\Part{Data:}
We use the \sbabi \cite{sestili2018sbabi} synthetic datasets for the \vd task.
The \sbabi dataset contains syntactically valid C programs with non-trivial control flow and buffer overflow vulnerability. It has been generated (and normalized) by Suneja \etal \cite{suneja2021probing} with \sbabi generator that generates almost $475$K functions. Samples with an `UNSAFE' tag (with synthetic vulnerability) are labeled as ``1'', while the ones with a `SAFE' tag (without vulnerability) are labeled as ``0''.

\Part{Models:}
We use two models for the \vd task: \rnn \cite{cho2014rnn} and \cnn \cite{kim2014cnn}. These models use different source code representations and architectures \cite{suneja2021probing}. The first model considers the source code as a sequence of tokens with a 2-layer bi-directional GRU, and the second model encodes the source code as an image with a 2d-convolutional layer.

\subsection{Method Name Prediction (\mn)}

\Part{Task:}
In the method name prediction task, the model attempts to predict the name of a method from its body. This task has been used as the downstream task to evaluate several state-of-the-art \ci models \cite{alon2019code2vec,alon2019code2seq}.

\Part{Data:}
We use the \jlarge dataset \cite{alon2019code2seq} for the \mn task which has about $16$M Java methods from $9500$ top-starred GitHub repositories. This dataset partitions $9000$ Java projects for training, $250$ Java projects for validation, and $300$ Java projects for testing.

\Part{Models}:
We use two models for the \mn task: \ctv \cite{alon2019code2vec}, and \cts \cite{alon2019code2seq}. These models rely on extracting ``paths'' from the method's abstract syntax tree (AST) that connect one terminal or token to another. \cts uses LSTMs to encode paths and decode targets one-by-one rather than un-tokenized embeddings as in \ctv.

\subsection{Variable Misuse Localization-Repair (\vm)}

\Part{Task:}
A variable misuse is a common bug in software development that occurs when a different variable is used than the intended variable in a program \cite{allamanis2018ggnn}. In the variable misuse localization and repair task, a model attempts to locate the misuse bug and propose a repair in the form of the correct identifier to use \cite{vasic2019neural, hellendoorn2020global}.

\Part{Data:}
We use the \pyg synthetic dataset derived by \citet{hellendoorn2020global} from the ETH Py150 dataset \cite{raychev2016tree}. This dataset contains functions from a total of 150K Python files - 90K files as the training set, 10K files as the validation set, and 50K files as the testing set. Each function is included both as a bug-free sample, and with up to three synthetically introduced bugs, yielding about 2 million samples in total.

\Part{Models:}
We use two models for the \vm task: \rnn \cite{cho2014rnn} and \tra \cite{vaswani2017attention}. The first model is a simple bi-directional recurrent-based architecture that uses GRU as the recurrent cell, and the \tra model is the attention-based architecture in which the representation of tokens is iteratively refined through all-to-all communication. We use 2 layers \rnn with 512 hidden dimensions, and 6 layers \tra with 8 heads and 512 attention dimensions.

\subsection{Natural Language Code Search (\cs)}

\Part{Task:}
In this task, given a natural language query, the target is to find the most semantically relevant source code from a collection of candidates. The task is formulated as a binary classification problem, where given a pair of query and code, a model aims to classify whether the code is semantically related to the query or not \cite{feng2020codebert}. It has been actively studied and applied in many software development practices \cite{CodeSearchNet, CodeXGLUE}.

\Part{Data:}
We use the preprocessed dataset derived by \citet{feng2020codebert} from the original \csn (CSN) dataset \cite{CodeSearchNet}, where each sample includes a code snippet paired with a natural language query. The dataset consists of a balanced number of positive and negative samples. Positive samples are the samples where the code is related to the query and are labeled as ``1''. Contrary, negative samples contain randomly replaced irrelevant code or query and are labeled as ``0''. We choose the Java language data (\csnjava) which contains more than $908$K samples for training, $30$K samples for validation, and $155$K samples for testing.

\Part{Models:}
The \cbert is a \texttt{bimodal} pre-trained model for programming languages (PL) and natural languages (NL). It captures the semantic connection between NL and PL and produces general-purpose representations that can broadly support various downstream NL-PL tasks \cite{feng2020codebert}. It has been developed following the architecture of BERT \cite{devlin2019bert} and RoBERTa \cite{liu2020roberta}, which itself is based on the Transformer {\cite{vaswani2017attention}} that is used in most large pre-trained models.

\begin{table*}
    \caption{Average probability change over samples.}
    \label{table:summary_probability_change}
    \def\arraystretch{1.2}
    
    \centering
    \resizebox{0.98\textwidth}{!}{%
    \begin{tabular}{|c|c|c|C{.8cm}|C{.8cm}|C{.8cm}|C{.8cm}|C{.8cm}|C{.8cm}|C{0.8cm}|C{0.8cm}|c|}
        \hline
        \multirow{2}{*}{\textbf{Task}} & \multirow{2}{*}{\textbf{Dataset}} & \multirow{2}{*}{\textbf{Model}}
        & \multicolumn{3}{c|}{Probability Increase (PI)}
        & \multicolumn{3}{c|}{Probability Decrease (PD)}
        & \multicolumn{3}{c|}{Sample ($\%$)}
        \\ \cline{4-12}
        & & 
        & \texttt{min} & \texttt{max} & \texttt{mean}
        & \texttt{min} & \texttt{max} & \texttt{mean}
        & PI & PD & PI $\cup$ PD
        \\ \hline 
        \hline
        
        \multirow{2}{*}{\vd} & \multirow{2}{*}{\sbabi} & \rnn 
        &  0.10 & 0.48 & 0.20
        & -0.10 & -0.48 & -0.19
        & 42.10 & 43.50 & 46.10
        \\ \cline{3-12}
        
        & & \cnn 
        & 0.10 & 0.49 & 0.23
        & -0.10 & -0.50 & -0.24
        & 32.20 & 54.00 & 54.70
        \\ \hline
        
        \multirow{2}{*}{\mn} & \multirow{2}{*}{\jlarge} & \ctv 
        & 0.10 & \textbf{0.72} & \textbf{0.25}
        & -0.10 & \textbf{-0.74} & \textbf{-0.27}
        & 46.20 & 64.10 & 65.20
        \\ \cline{3-12}
        
        & & \cts 
        & 0.10 & 0.55 & 0.18
        & -0.10 & -0.69 & -0.22
        & 44.90 & 72.70 & 74.80
        \\ \hline
        
        \multirow{2}{*}{\vm} & \multirow{2}{*}{\pyg} & \rnn 
        & 0.10 & 0.45 & 0.22
        & -0.10 & -0.64 & -0.24
        & 85.00 & 93.50 & 94.40
        \\ \cline{3-12}
        
        & & \tra 
        & 0.10 & 0.59 & 0.22
        & -0.10 & -0.60 & -0.23
        & 79.80 & 95.30 & 95.60
        \\ \hline

        \cs & \csnjava & \cbert
        & 0.10 & 0.46 & 0.19
        & -0.10 & -0.47 & -0.21
        & 28.50 & 34.40 & 46.90
        \\ \hline
        
    \end{tabular}%
    }
\end{table*}

\section{Results and Discussion}

We have used models and datasets of the four tasks from the literature \cite{rabin2021dd, suneja2021probing, CodeXGLUE} and randomly picked $1,000$ sample input programs from the test set for our analysis.
We study the intermediate reduced programs that both positively and negatively contribute to the prediction of the model.
By finding and understanding tokens that affect the confidence of models, we hope to gain insights that can be useful to debug the model's predictions.
In this section, we quantify the effect of distractors on the model's confidence across various tasks, models, and datasets.
We summarize the probability changes caused by distractors and highlight the top distractor tokens with categories.
This type of analysis is an important step in leading models to improve transparency.

\subsection{RQ1: Effect of distractors}

Towards observing the effect of distractors on tasks and models, we study the percentage of samples that yielded a noticeable probability change, at least $10\%$. We compute the average value of this change along with min and max change, in both directions (increasing and decreasing), during the reduction steps -- all of which are summarized in Table~\ref{table:summary_probability_change}.

From the `Sample (\%)' of \Cref{table:summary_probability_change}, we can see that a probability increase (PI) is reported on up to $42\%$ of \sbabi samples (by \rnn model), $46\%$ of \jlarge samples (by \ctv model), $85\%$ of \pyg samples (by \rnn model), and $28\%$ of \csnjava samples (by \cbert model). On the other hand, a probability decrease (PD) is reported on up to $54\%$ of \sbabi samples (by \cnn model), $72\%$ of \jlarge samples (by \cts model), $95\%$ of \pyg samples (by \tra model), and $34\%$ of \csnjava samples (by \cbert model).
These results show that \textbf{the \vm task (the \rnn and \tra models on the \pyg dataset) exhibit more distractor cases on the test samples}, which is almost $20\%$, $40\%$, and $50\%$ higher than the \mn, \vd, and \cs tasks, respectively. Note that the maximum possible reduction in the \vm task is limited because it needs to preserve all variable occurrences as potential error and repair targets \cite{rabin2021dd}, thus, removing tokens may result in a noticeable probability change over $90\%$ of samples.

On comparing the models' PI and PD values from \Cref{table:summary_probability_change}, we additionally observe that the \ctv model has the highest `max' (PI=0.72, PD=-0.74) and `mean' (PI=0.25, PD=-0.27) probability change values compared to other models.
The other models' PI and PD values where are more ambivalent -- \eg, the \cts model has a relatively low average PI value (=0.18) on the \jlarge samples, and the \rnn model has a relatively low average PD value (=-0.19) on the \sbabi samples.
These findings may suggest, \textbf{on average, the \ctv model is comparatively more susceptible to distractors than the other models}.

\begin{figure*}
    \centering
    \begin{minipage}{.48\textwidth}
        \centering
        \includegraphics[width=\linewidth]{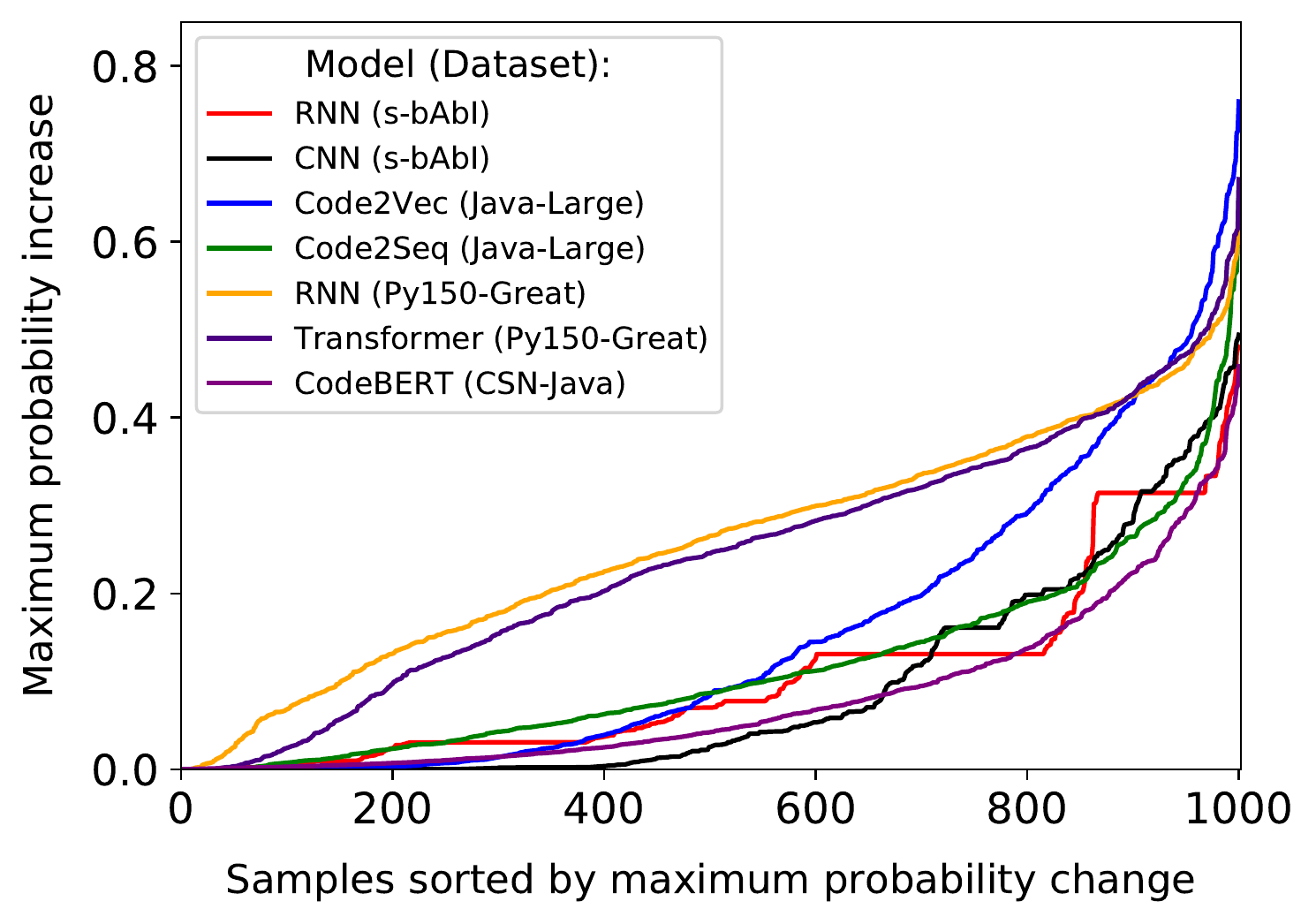}
        \caption*{(a) Maximum Probability Increase (MPI)}
    \end{minipage}%
    \begin{minipage}{0.48\textwidth}
        \centering
        \includegraphics[width=\linewidth]{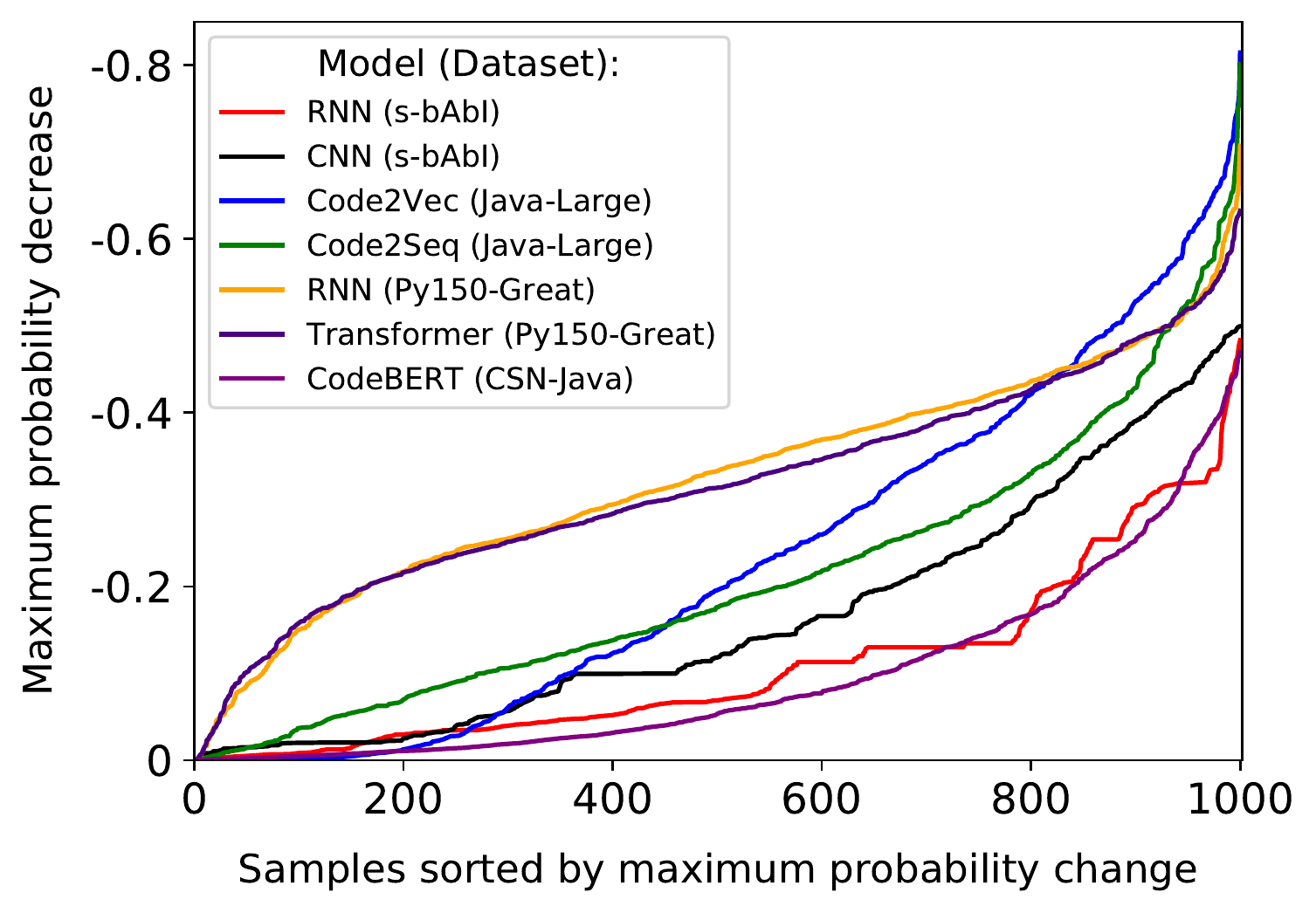}
        \caption*{(b) Maximum Probability Decrease (MPD)}
    \end{minipage}
    \caption{Maximum probability change over samples.}
    \label{fig:max_probability_change}
\end{figure*}

\subsection{RQ2: Extent of probability change}

To understand the extent of distractors where the model's confidence changes significantly, we study the maximum increase and decrease of a model's prediction probability score in this section. \Cref{fig:max_probability_change} shows the maximum probability change by models that were reported at any step in the reduction process of samples. \Cref{fig:max_probability_change}(a) highlights the maximum probability increase (MPI) and \Cref{fig:max_probability_change}(b) shows the maximum probability decrease (MPD).

According to \Cref{fig:max_probability_change}, the \rnn(\pyg) and \tra(\pyg) models of \vm task have higher MPI and MPD values for the majority of samples than those reported by the models of other tasks. The \cbert(\csnjava) model of \cs task happens to report the lowest MPI and MPD values. For the \vd task, the \rnn(\sbabi) model shows lower MPD while the \cnn(\sbabi) model shows lower MPI. The models of \mn task (\ctv and \cts) have trended in the middle for both MPI and MPD. These results may indicate that \textbf{the \cbert model for \cs task has a comparatively higher reliance on individual tokens than other models}, in our experiments.


\begin{table}
    \caption{Top distractor tokens where model's prediction probability score changes by at least $10\%$.}
    \label{table:top_reduction_tokens}
    \def\arraystretch{1.2}
    
    \centering
    \resizebox{0.98\columnwidth}{!}{%
    \begin{tabular}{|c|c|l|}
        \hline
        \textbf{Task} & \multicolumn{2}{c|}{\textbf{Top Distractor Tokens}} \\ \cline{2-3}
        (Dataset) & \textbf{Categories} & \textbf{List of Tokens} \\
        \hline
        \hline

        & control-flow & if, else, for, while, return \\ \cline{2-3}
        & datatypes & int, char \\ \cline{2-3}
        \vd & digits & NUMBER, 0, 1 \\ \cline{2-3}
        (\sbabi) & identifiers & VARIABLE, rand \\ \cline{2-3}
        & operators & =, $<$, ++ \\ \cline{2-3}
        & others & ; \\ \hline

        & control-flow & if, for, return \\ \cline{2-3}
        & datatypes & None \\ \cline{2-3}
        \vm & digits & 0, 1 \\ \cline{2-3}
        (\pyg) & identifiers & assertEqual \\ \cline{2-3}
        & operators & ., =, ==, :, in, $\%$ \\ \cline{2-3}
        & others & NEWLINE, INDENT, UNIND \\ \hline
        
        & control-flow & if, return \\ \cline{2-3}
        & datatypes & String \\ \cline{2-3}
        \mn & digits & 0, 1 \\ \cline{2-3}
        (\jlarge) & identifiers & qname \\ \cline{2-3}
        & operators & ., =, !, : \\ \cline{2-3}
        & others & ;, @, Override, super, new, this \\ \hline
        
        & control-flow & if, return, try \\ \cline{2-3}
        \cs & datatypes & String, int \\ \cline{2-3}
        (\csnjava) & modifiers & public, final, static \\ \cline{2-3}
        & operators & ., =, $<$, $>$ \\ \cline{2-3}
        & others & ;, @, Override, new\\ \hline
        
    \end{tabular}%
    }
\end{table}

\subsection{RQ3: Top distractor tokens}
 
We next study the category of distractor tokens that are removed from input programs in reduction steps, especially when the probability score changes significantly.
Table~\ref{table:top_reduction_tokens} shows such most popular tokens that are removed from input programs and contributed to at least a $10\%$ change in the probability score of a model's prediction.
We have found a total of $16$ unique tokens that are distractor tokens in the normalized \sbabi dataset considering all models. Therefore, we are only highlighting the top $16$ common distractor tokens among models across different tasks and datasets.

From Table~\ref{table:top_reduction_tokens}, we can observe the distractor tokens to be from the following categories: control-flow (\eg, `if', `else', `for', `while', `return', `try'), datatypes (\eg, `int', `char', `String'), digits (\eg, `0', `1', `numbers'), operators (\eg, `.', `=', `$<$', `$>$', `==', `++', `\%'), end-of-statement syntax (\eg, `;', `NEWLINE'), and other identifiers and keywords.
Interestingly, we can see that tokens from these categories are among the most popular distractor tokens for all four prediction tasks. Thus, models of all tasks seemed to be impacted (at least $10\%$ in prediction probability score) by the similar categories of tokens (note that, there are no datatypes for the \vm task, which works on Python code). 
These results may highlight that \textbf{tokens from a variety of categories can play a significant role in the model's confidence}. 
This finding merits further analysis of programs where these categories can also be part of the important tokens that are actually responsible for predictions. We look forward to providing a comparative study in our future work.
 
\section{Related Work}
\label{sec:related}

There has been a lot of work in the area of transparent or interpretable machine learning for computer vision, text, and natural language processing, that focuses on understanding the underlying reasoning of neural models \cite{samek2020interpretable, rauker2022transparent}.
Interpretable or transparent machine learning has numerous benefits for \ci, including making predictions explainable \cite{rabin2021dd, wang2022demystifying}, identifying reasoning about mispredictions \cite{suneja2021probing, cito2022counterfactual}, understanding key features \cite{bui2019autofocus, rabin2022perses}, using learned models to generate new insights \cite{rabin2021generalizability, rabin2023memorization}, and improving the quality of the models themselves \cite{suneja2021enhancement, zhang2022dietcode, wang2022bridging}.

\subsection{Learning Code Elements}
Models often learn specific features, simple shortcuts, or even noise for achieving target performance.
\citet{compton2020obfuscation} show that the code2vec embeddings highly rely on variable names and investigate the effect of obfuscation on improving code2vec embeddings that better preserve code semantics.
Following the generalizability of word embeddings, \citet{kang2019generalizability} assess the generalizability of code embeddings in various software engineering tasks and demonstrate that the learned embeddings by code2vec do not always generalize to other tasks beyond the example task it has been trained for.
\citet{rabin2019tnpa,rabin2020evaluation,rabin2021generalizability} and \citet{yefet2020adversarial} demonstrate that the models of code often suffer from a lack of robustness or generalizability and are vulnerable to adversarial examples.
\citet{suneja2021probing,suneja2021enhancement} uncover the model’s reliance on incorrect signals by checking whether the vulnerability in the original code is missing in the reduced minimal snippet.
\citet{rabin2021dd,rabin2022perses} demonstrates that models often use just a few simple syntactic shortcuts for making predictions.
\citet{allamanis2019duplication} find that deep learners are easily led astray by factors like code duplication and this repetition spuriously inflated their performance.
\citet{rabin2023memorization} later show that models can fit noisy training data with excessive parameter capacity and thus suffer in generalization performance.
As models often learn noise or irrelevant features for achieving high prediction performance, the lack of understanding of what input features impact models' prediction would hinder the trustworthiness to correct classification. Such opacity is substantially more problematic in critical applications such as vulnerability detection or automated defect repair. Thus, we try to highlight distractor tokens for improving transparency.

\subsection{Extracting Key Features}
Several kinds of research have been done in finding relevant input features for models of source code.
\citet{allamanis2015suggesting} use a set of features from programs and show that extracting relevant features that capture global context is essential for learning effective code context. \citet{rabin2020demystifying} attempt to find key input features of a label by manually inspecting some input programs of that label.
\citet{bui2019autofocus} attempt to identify relevant code elements by perturbing statements of the program and combining corresponding attention and confidence scores.
\citet{cito2022counterfactual} integrate minimal changes to code to find realistic and plausible counterfactual explanations under which a model changes its prediction.
\citet{wang2022demystifying} propose a mutate-reduce approach to find key features in the code summarization models considering valid programs.
\citet{rabin2021dd,rabin2022perses} and \citet{suneja2021probing,suneja2021enhancement} use input program reduction techniques to find minimal inputs that preserve the model's prediction, hence finding key tokens in the program with respect to the prediction.
By removing irrelevant parts to a prediction from the input programs, the authors aim to better understand key features in the model inference.
We base our approach on this \DD algorithm and study the impact of tokens after removing them from input programs.

\section{Conclusion}
\label{sec:conclusion}

In this paper, we introduced the notion of distractor features in \ci models.
We proposed a technique based on \DD to study the features that impact the confidence of the model in its prediction. 
We evaluate our approach across several tasks, models, and datasets of code.
We observe that the \ctv model is more likely to be susceptible to distractors while the \cbert model has a comparatively higher reliance on individual tokens.
In the future, we plan to conduct a detailed study on the impact and extent of distractors in various large generative models considering different aspects of training and testing.

\balance
\bibliography{refs}
\bibliographystyle{IEEEtranN}
\end{document}